\documentclass[10pt,twocolumn,letterpaper]{article}

\usepackage{cvpr}
\usepackage{times}
\usepackage{epsfig}
\usepackage{graphicx}
\usepackage{caption}
\usepackage{amsmath}
\usepackage{amssymb}
\usepackage{bm}
\usepackage{algorithm}
\usepackage{algpseudocode}
\usepackage{multirow}
\usepackage[tight,footnotesize]{subfigure}

\algnewcommand\algorithmicforeach{\textbf{for each}}
\algdef{S}[FOR]{ForEach}[1]{\algorithmicforeach\ #1\ \algorithmicdo}

\newcommand{\Tref}[1]{Table~\ref{#1}}
\newcommand{\eref}[1]{Eq.~(\ref{#1})}

\newcommand{\fref}[1]{Fig.~\ref{#1}}
\newcommand{\Fref}[1]{Figure~\ref{#1}}
\newcommand{\sref}[1]{Sec.~\ref{#1}}

\newcommand{\vx}{\mathbf{x}}
\newcommand{\vp}{\mathbf{p}}
\newcommand{\vr}{\mathbf{r}}
\newcommand{\vF}{\mathbf{F}}
\newcommand{\vg}{\mathbf{g}}
\newcommand{\vt}{\mathbf{t}}

\usepackage[breaklinks=true,bookmarks=false,hyperfootnotes=false]{hyperref}

\newcommand\blfootnote[1]{%
  \begingroup
  \renewcommand\thefootnote{}\footnote{#1}%
  \addtocounter{footnote}{-1}%
  \endgroup  
}

\cvprfinalcopy 

\def\cvprPaperID{1443} 

\makeatletter
  \let\ps@plain\ps@empty
\makeatother

\ifcvprfinal\fi
\begin{document}

\title{Probabilistic Plant Modeling via Multi-View Image-to-Image Translation}

\author{
Takahiro Isokane$^{1,*}$
\quad
Fumio Okura$^{1,2,*}$
\quad
Ayaka Ide$^1$
\quad
Yasuyuki Matsushita$^1$
\quad
Yasushi Yagi$^1$\\
$^1$Osaka University
\qquad 
$^2$JST PRESTO\\
{\tt\small \{isokane,okura,ide,yagi\}@am.sanken.osaka-u.ac.jp \qquad yasumat@ist.osaka-u.ac.jp}
}



\twocolumn[{%
\renewcommand\twocolumn[1][]{#1}%

\maketitle

\begin{center}
	\centering
	\includegraphics[width=\linewidth]{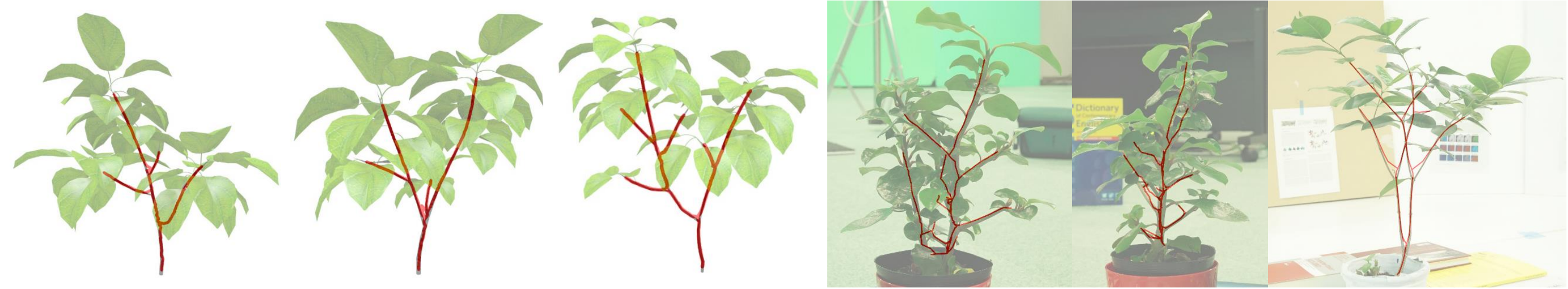}  
    \captionof{figure}{Result of branch structure estimation from simulated and real leafy plant images. From multi-view plant images, our method infers the branch structure in a probabilistic framework. An explicit graph structure (red lines) can be derived from the probabilistic plant model. }
	\label{fig:teaser}
\end{center}
\vspace{1mm}
}]

\begin{abstract}
This paper describes a method for inferring three-dimensional (3D) plant branch structures that are hidden under leaves from multi-view observations. Unlike previous geometric approaches that heavily rely on the visibility of the branches or use parametric branching models, our method makes statistical inferences of branch structures in a probabilistic framework. By inferring the probability of branch existence using a Bayesian extension of image-to-image translation applied to each of multi-view images, our method generates a probabilistic plant 3D model, which represents the 3D branching pattern that cannot be directly observed. Experiments demonstrate the usefulness of the proposed approach in generating convincing branch structures in comparison to prior approaches.
\end{abstract}

\blfootnote{$^*$Authors contributed equally.}
\vspace{-1em}

\section{Introduction}
\vspace{-1mm}
We propose an approach to estimate branch (skeleton) structures of plants from multi-view images that are severely occluded by leaves. Unveiling hidden skeleton structures is one of the most challenging tasks in computer vision, because it naturally involves inference of the unobserved structures. Skeleton estimation is actively studied for human pose estimation~\cite{cao17}. A recent trend in image-based human skeleton estimation relies on the prior knowledge of the relationship among joints for extracting and interpreting joints from images. Unlike human bodies, the branch structures of plants are less organized, namely, the number of joints and their connections are generally unknown. In addition, heavy occlusions due to leaves makes the problem harder. These aspects pose a unique challenge in plant's structure estimation. Although difficult, branch structure estimation has a number of applications, such as synthetic plant generation in computer graphics and computational agriculture.

Toward this goal, this paper presents a multi-view image-to-image translation approach to 3D branch structure estimation. To estimate the branching paths hidden under leaves, we cast the estimation problem to an \emph{image-to-image translation} problem~\cite{isola17}, which converts an image from one domain to another domain. In our context, we convert an input image of a leafy plant to a map that represents the branch structure, \ie, each pixel containing the prediction of ``branch'' or ``not-branch''. To deal with the uncertainty of the prediction, we develop a Bayesian extension of image-to-image translation applied to each of the multi-view images. It yields the prediction together with its reliability, resulting in \emph{probabilistic} estimates in a similar manner to a Bayesian semantic segmentation approach~\cite{kendall17}.

The probabilistic representation is advantageous in a few important aspects; not only that we have access to the credibility of the estimates, but also that it allows us to consolidate the view-dependent inferences in a 3D space in a well-defined probabilistic framework. Thus, in our method, instead of directly predicting 3D branch structures, it makes inference in individual views of the multi-view input and subsequently aggregate them in a 3D voxel space to obtain a probabilistic 3D plant model. We also develop a method for generating explicit branch structures from the probabilistic model based on particle flows so that the probabilistic representation can be converted to a form that could be used in intended applications. A few examples of the output are shown in \fref{fig:teaser}.

%
The primary contributions of this study are twofold. First, we propose a method of recovering 3D structures of a plant based on Bayesian image-to-image (leafy- to branch-image) translation applied in a multi-view manner.
By representing the \emph{branchness} in a probabilistic framework, we show that both faithful estimates and comprehensive aggregation over multi-view predictions can be achieved. Second, we show that the branch structure estimation is made possible by the proposed approach, together with a method for extracting explicit branch structures from the probabilistic representation. Experimental results show that the proposed method can generate convincing branch structures even with severe occlusions by leaves and other branches. In comparison to a traditional tree reconstruction approach, we found notable improvement in the resultant 3D branching structure. 



\section{Related Works}
\vspace{-1mm}
Our goal is to reconstruct a 3D branch structure of a plant from multi-view images that exhibit severe occlusions due to plant leaves. At the heart of the proposed method, we develop a Bayesian extension of image-to-image translation for making statistical predictions of branch existence even for the parts that are not observed at all. In what follows, we discuss the prior arts for 3D reconstruction of plants and trees, image-to-image translation and Bayesian neural networks that are related to our work.

\vspace{-4mm}
\paragraph{3D reconstruction of plants and trees.}
(Semi-) automatic 3D modeling of plants and trees is actively studied in the graphics community~\cite{waite88} because of their importance as a rendering subject and that their manual modeling is notably time-consuming. Interactive 3D modeling methods of trees using manually provided hints, \eg, lines, are proposed in the early 00's~\cite{boudon03,okabe05}. Growth models of trees, which include branching rules, are occasionally utilized~\cite{galbraith04,streit05,palubicki09}, some of them are augmented by realistic textures~\cite{livny11}. These approaches generally heavily rely on the branch models (or rules), and the resulting structure cannot deviate much from the presumed models.

On the other hands, an approach based on observations of real-world trees that uses photographs~\cite{shlyakhter01} or 3D scans~\cite{xu07,livny10} is shown promising for automatic tree modeling. Several tree modeling approaches using multi-view images have been proposed~\cite{reche04,tan07,neubert07}.
These approaches have been further extended to single-image based methods~\cite{tan08,argudo16} for better applicability.
The major focus of the most image-based modeling approaches has been to generate 3D tree models that well fit the silhouette or volumes, not necessarily aiming at recovering branch structures. 
Using multi-view images of bare trees, previous approaches achieve geometric reconstruction of a branch structure~\cite{lopez10,stava14,zhang15}; however, leaves of plants or trees make 3D reconstruction considerably difficult due to occlusions. 
The scope of plant modeling now goes beyond computer graphics and is becoming an important application of computer vision, \ie, analysis of plant shapes and growth for vision-assisted cultivation and plant phenotyping. Our method aims to recover branch structures via a multi-view approach with an emphasis on dealing with uncertainty due to severe occlusions by leaves.

Apart from the plant and tree context but somehow related, recently human skeleton (structure) estimation shows a great success~\cite{pishchulin16,cao17}. The task has a similarity to what we study in this paper; however, the plant and tree branch structure estimation has its unique difficulty in that the number of joints and their relationships are generally unknown and complex. 

\begin{figure*}
\begin{center}
\includegraphics[width=170mm]{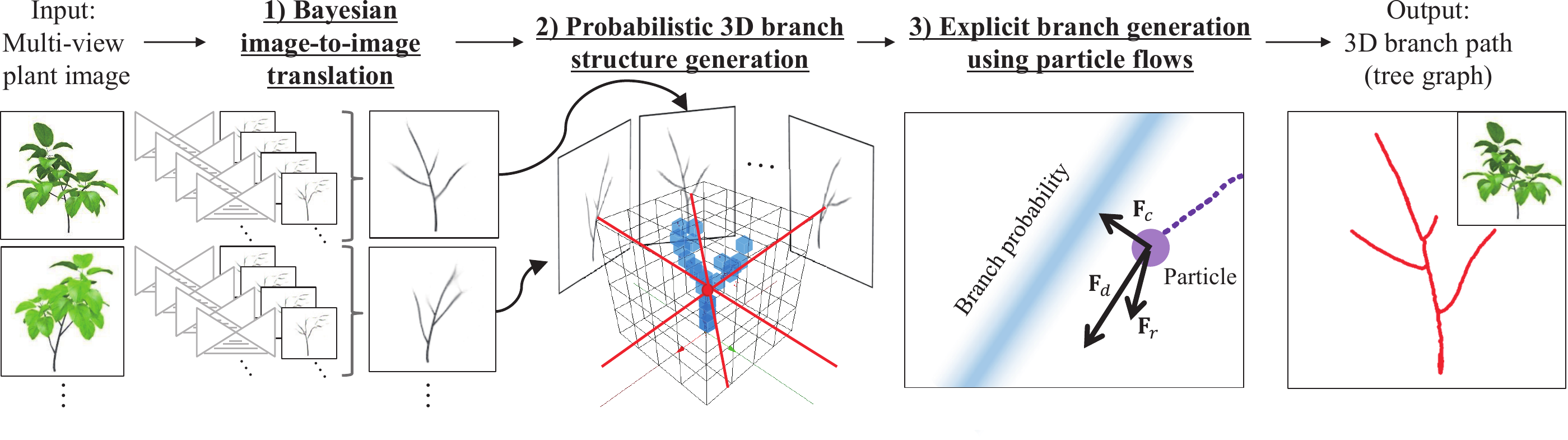}
\end{center}
\vspace{-5mm}
   \caption{{\bf Overview of our plant's branch structure modeling.} The proposed method first estimates the 2D branch existence probability on multi-view images using a Bayesian extension of image-to-image translation. The 2D probabilities are then consolidated in a 3D voxel space to form a volume of 3D branch probability. An explicit 3D branch structure is generated by particle flow simulation based on the 3D probability map.}
\label{fig:overview}
\end{figure*}

\vspace{-4mm}
\paragraph{Image-to-image translation.}
Image-to-image translation aims at transferring contextual or physical variation between the source and target images. The early works on image-to-image translation include image analogies~\cite{hertzmann01} and texture transfer~\cite{efros01}. Commonly, these approaches divide the image into small patches and transfer the change based on patch-wise correspondences~\cite{lefebvre05,barnes09,darabi12}.
More recent image-to-image translation largely benefits from deep learning, such as convolutional neural networks (CNNs) with encoder-decoder architectures or using Generative Adversarial Networks (GANs). Pix2Pix~\cite{isola17}, which uses conditional GAN, shows impressive performance on a wide variety of translation tasks. Along this context, Cycle GAN~\cite{zhu17} has shown the possibility of image-to-image translation without paired training images. 

\vspace{-4mm}
\paragraph{Bayesian neural networks}
To obtain \emph{reliability} of inference, a Bayesian framework has been used together with neural networks in Bayesian neural networks (BNNs)~\cite{neal12}. 
Recent BNNs use dropout connections in forward passes in the prediction stage~\cite{gal16}, which yields variations in inference. The estimate is thus given in the form of distribution rather than a point, effectively modeling the uncertainty of the prediction. This approach can be regarded as an approximation of Monte Carlo integration in traditional Bayesian approaches~\cite{li17}. An impressive result has been shown in Bayesian SegNet~\cite{kendall17} that achieves semantic segmentation augmented by the prediction reliability. Our method also uses the Bayesian approach in image-to-image translation task in order for obtaining the probabilistic representation of the branch structures.

%

\section{Probabilistic Branch Structure Modeling}
\vspace{-1mm}
Our method takes as input multi-view images of a plant and generates a probabilistic 3D branch structure in a 3D voxel space. Our method begins with estimating a 2D probabilistic branch existence map in each of the multi-view images based on an altered image-to-image translation method. Once the probabilistic branch existence map is computed for each view, they are merged in a 3D voxel space using the estimated camera poses based on a structure-from-motion method~\cite{visualsfm} to yield a probabilistic 3D branch structure. Finally, an explicit 3D branch structure is generated by a particle flow simulation, which is inspired by a traditional tree modeling approach~\cite{neubert07}, using the probabilistic 3D branch structure. \Fref{fig:overview} illustrates the whole pipeline of the proposed method. In what follows, we explain the individual steps of the proposed method.

\subsection{Bayesian image-to-image translation}
\vspace{-1mm}
\label{subsec:bayesian_branch_probability_generation}
From a leafy plant image, we first estimate a pixel-wise 2D branch existence probability. The major challenge is to infer branch structure hidden under leaves, which cannot be directly observed by a camera, possibly not from any of the viewpoints. It is here that we adopt a Pix2Pix approach~\cite{isola17} to image-to-image translation as a mean to derive a statistically valid prediction of the existence of branches in the multi-view images. For our context, we train a Pix2Pix network using pairs of a leafy plant and its corresponding label map describing the branch region.

\begin{figure}[t]
\begin{center}
\includegraphics[width=\linewidth]{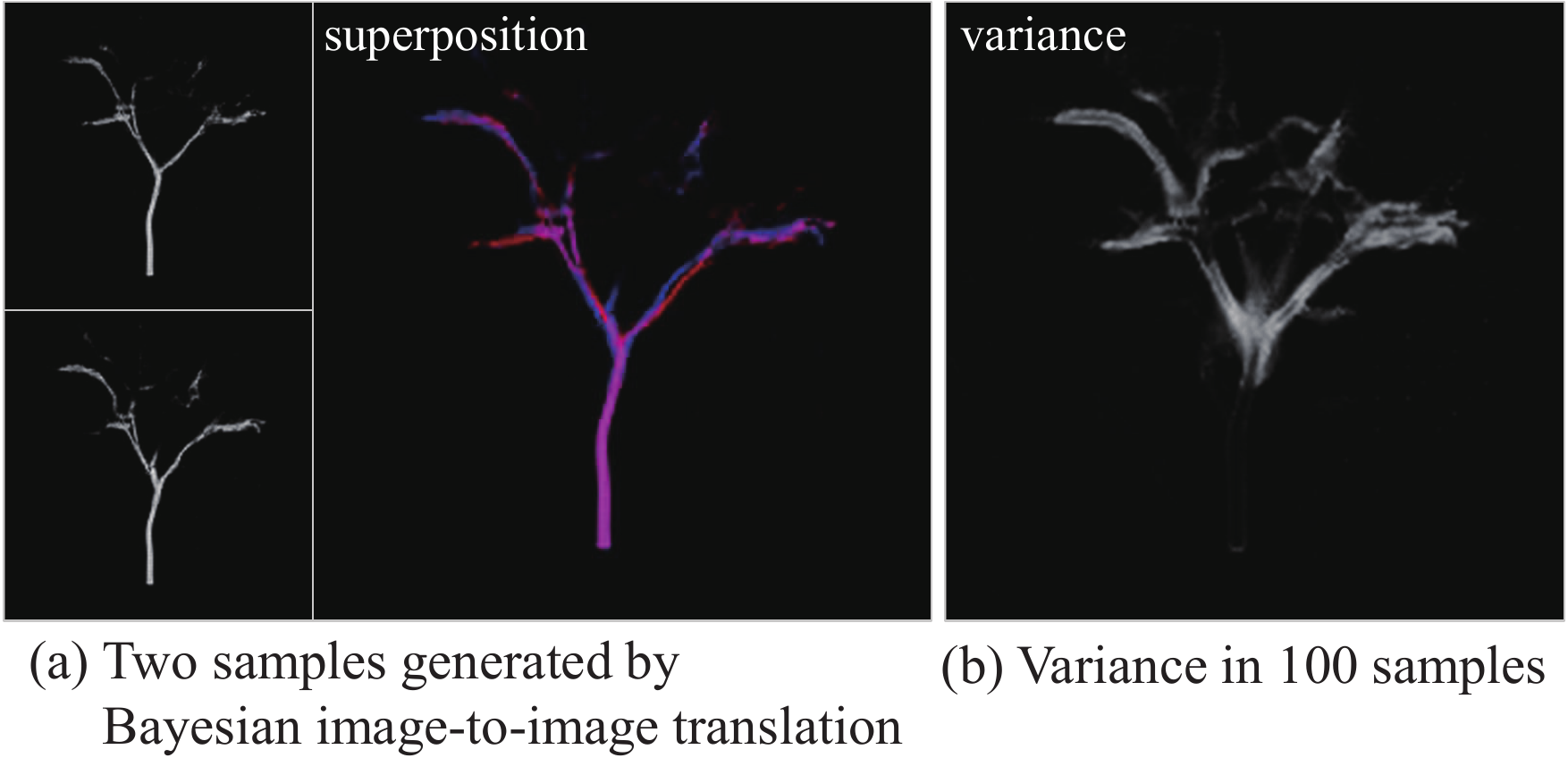}
\end{center}
\vspace{-3mm}
   \caption{Effect of Bayesian image-to-image translation in branch generation. Two generated samples contain differences in small branches (a). The variance in a larger number of samples (b) shows the uncertainty in the prediction, which cannot be obtained from a single inference.}
\label{fig:p2p}
\end{figure}

\begin{figure*}[t]
\begin{center}
\includegraphics[width=160mm]{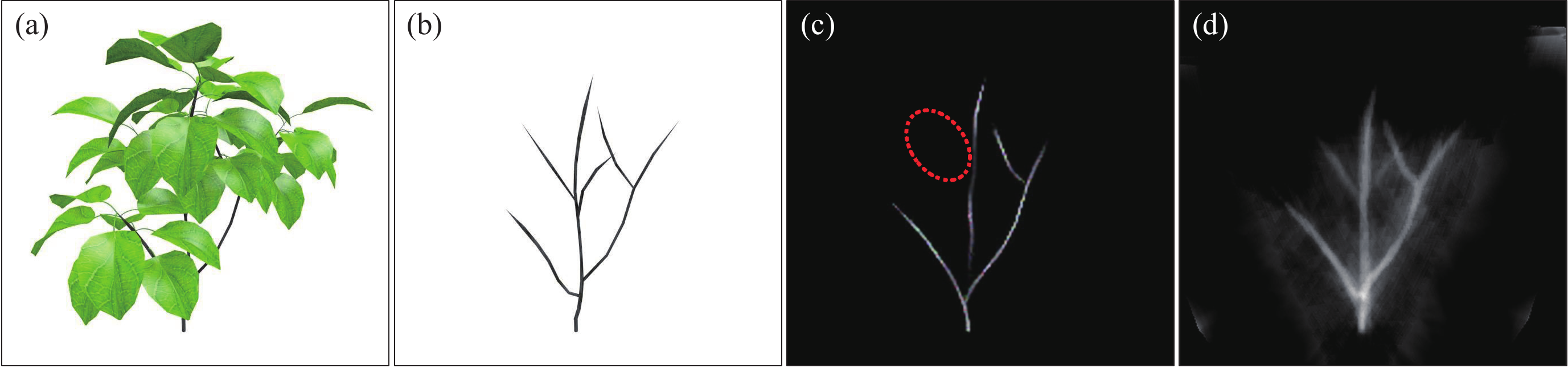}
\end{center}
\vspace{-2mm}
   \caption{Visualizations of branch probability: a) an input image, b) the ground truth branch, c) 2D branch probability generated by Bayesian image-to-image translation, d) a 2D projection of 3D branch probability via 3D aggregation. Although image-to-image translation (c) infers a few branches as low probability due to heavy occlusions (\eg, in the circle), they are recovered by 3D probability generation (d) due to votes from other views.}
\label{fig:prob}
\end{figure*}

To obtain a meaningful \emph{probabilistic} branch existence as output, we use the original image-to-image translation approach in a Bayesian deep learning framework~\cite{gal16, neal12} by Monte Carlo sampling via dropout at the inference stage. By adopting this fluctuation and repeating the prediction multiple times, we obtain a probability of the branch existence for each pixel. 
This strategy is implemented by inserting dropout layers before the middle four convolution layers in the encoder-decoder network of Pix2Pix. To further increase variations of inference, dropout is additionally applied to skip connections~\cite{ronneberger15} of Pix2Pix. \Fref{fig:p2p} shows an example of output variations by this Bayesian branch generation. Multiple inferences yield the degree of uncertainty, which cannot be obtained by a single inference.

As such, we obtain the mean of variational inferences from the image-to-image translation network. By treating the variational inferences as stochastic samples, each pixel in the mean inference can be regarded as the probability of branch existence spanned in $[0,1]$. For the $i$-th image $I_i$, the branch probability $B_{{2D}_i} : \mathbb{R}^2 \rightarrow [0, 1]$ at a pixel $\vx_{2D} \in \mathbb{R}^2$ is written as:
\begin{align}
\left\{
\begin{aligned}
& B_{{2D}_i}(\vx_{2D}) = \frac{1}{n_v} \sum_v \hat{B}_{{2D}_{i,v}}(\vx_{2D}), \\
& \hat{B}_{{2D}_{i,v}} = \pi_v (I_i), 
\end{aligned}
\right.
\label{eq:2D_branch_probability}
\end{align}
where $\pi_v$ denotes the Pix2Pix translation from an image $I_i$ to the corresponding branch existence $\hat{B}_{{2D}_{i,v}}$ with the $v$-th variation of random dropout patterns. The probability map $B_{{2D}_i}$ for each viewpoint $i$ is then obtained by marginalizing the individual samples $\hat{B}_{{2D}_{i,v}}$ over random trials $v$.

\subsection{Probabilistic 3D branch structure generation}
\vspace{-1mm}
Once the view-wise probability maps $\{B_{{2D}_i}\}$ are obtained, our method estimates a 3D probability map $B_{3D}$ of the branch structure defined in the 3D voxel coordinates. From the multi-view input images, we estimate the camera poses and intrinsic parameters by a structure-from-motion method~\cite{visualsfm}. 
It yields a set of projection functions $\{\theta\}$ that map from the 3D voxel to image coordinates, $\theta: \mathbb{R}^3 \rightarrow \mathbb{R}^2$. Using the projections $\{\theta\}$, the probability of the branch existence $B_{3D}$ at voxel $\vx_{3D} \in \mathbb{R}^3$ can be computed as a joint distribution of $\{B_{2D_{i}}\}$ by assuming their independence as
\begin{align}
B_{3D}(\vx_{3D}) = \prod_{i} B_{2D_{i}}(\theta_i(\vx_{3D})),
\label{eq:aggregation}
\end{align}
in which $\theta_i$ represents a projection from the voxel to the $i$-th image coordinates.
This aggregation process can be regarded as a back projection, which is used in traditional computed tomography~\cite{brooks75}. While any of the views may not convey complete information of the branch structure due to heavy occlusions, the aggregation effectively recovers the branch structure in a probabilistic framework as depicted in \fref{fig:prob}~(d). To avoid numerical instability, \eref{eq:aggregation} is computed in the log domain.

\subsection{Explicit branch generation using particle flows}
\begin{figure}[t]
\begin{center}
\includegraphics[width=\linewidth]{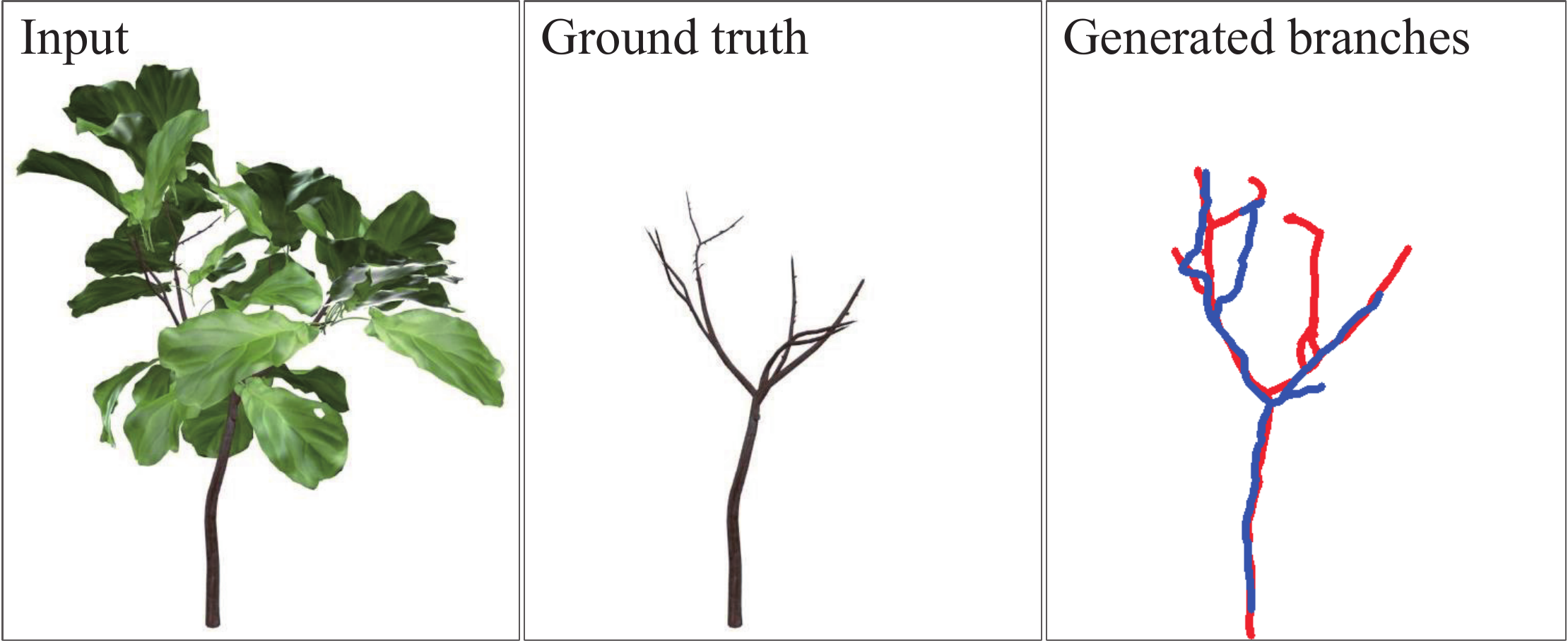}
\vspace{-7mm}
\end{center}
   \caption{Different branch structures (shown in red and blue) generated from the same branch probability map, by changing position and the number of initial particles.}
\label{fig:diff_structures}
\vspace{-1mm}
\end{figure}

\vspace{-1mm}
\label{subsec:paticle_flow}
The probabilistic 3D branch structure can be converted to an explicit representation of 3D branch models that can be used for applications in computer graphics and branch structure analysis. Inspired by a conventional tree modeling approach~\cite{neubert07}, we develop a branch structure generation method using particle flows. Instead of relying on the attractor graph that is computed directly from images in~\cite{neubert07}, our method uses the 3D probability map to regulate the particle movement for generating branch path candidates. Finally, the candidates are consolidated via structure refinement that includes smoothing and simplification. The resulting 3D model is represented by a graph that consists of nodes and edges that correspond to joints and branches.
Using the particle-flow-based method, our method can yield different plant structures from a single probability map, by changing position and numbers of initial particles as shown in \fref{fig:diff_structures}. This is one of the merits of our probabilistic plant modeling for graphics applications.

\begin{figure}[t]
\begin{center}
\includegraphics[width=\linewidth]{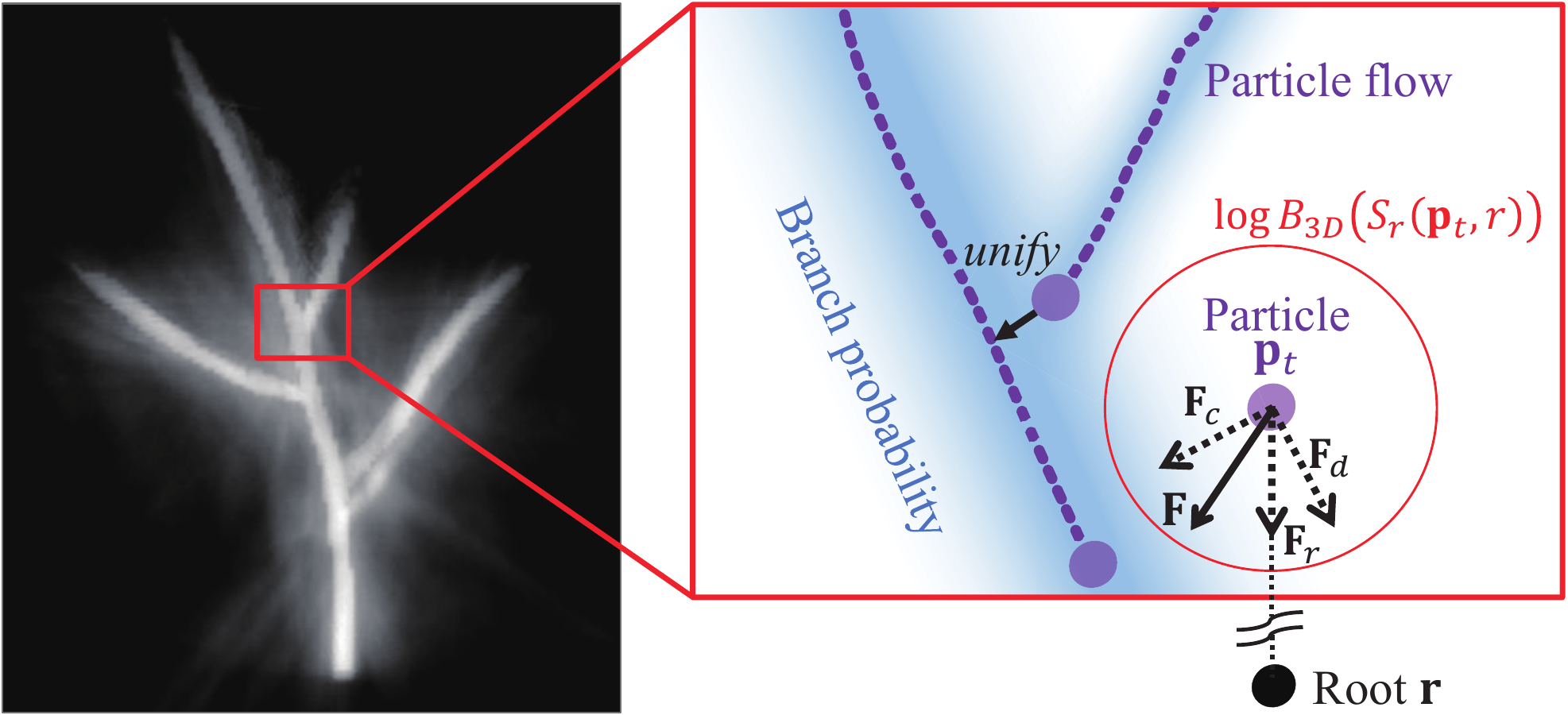}
\vspace{-7mm}
\end{center}
   \caption{Particle flow simulation. At each step, particles move according to the force computed from the 3D branch probability. The traces of particles are stored as branch path candidates.}
\label{fig:particle}
\end{figure}

We first generate particles proportionally to the log probability map $\log B_{3D}$ for avoiding the peaky distribution in the canonical domain. Also, the root position $\vr \in \mathbb{R}^3$ of the plant is set to the bottommost point that has a high probability of being a part of branches. Under these settings, starting from random distribution of particles, the particle positions are iteratively updated in a manner of flow simulation. At each step $t$, the $(t+1)$-th particle position $\vp_{t+1}$ is computed by the following update rules:
\begin{align}
\left\{
\begin{aligned}
\vp_{t+1}    &\leftarrow \vp_{t} + \vF(\vp_{t}), \\
\vF(\vp_{t}) &= \lambda_c \vF_c(\vp_{t}) +\lambda_d \vF_d(\vp_{t}) + \lambda_r \vF_r(\vp_{t}).
\end{aligned}
\right.
\end{align}
As illustrated in \fref{fig:particle}, $\vF_c(\vp_{t})$ and $\vF_d(\vp_{t})$ represent normalized vectors toward and parallel to the stream of branch probability. $\vF_r(\vp_{t})$ represents the unit direction from $\vp_{t}$ to the root point $\vr$ of the plant. These directions are linearly combined with weight factors $\lambda_c$, $\lambda_d$, and $\lambda_r$ that are determined empirically.
The traces of the particles are recorded in a tree graph as vertices and edges, and a unification of particles is treated as a joint.

\begin{figure}[t]
\begin{center}
\centering
\includegraphics[width=\linewidth]{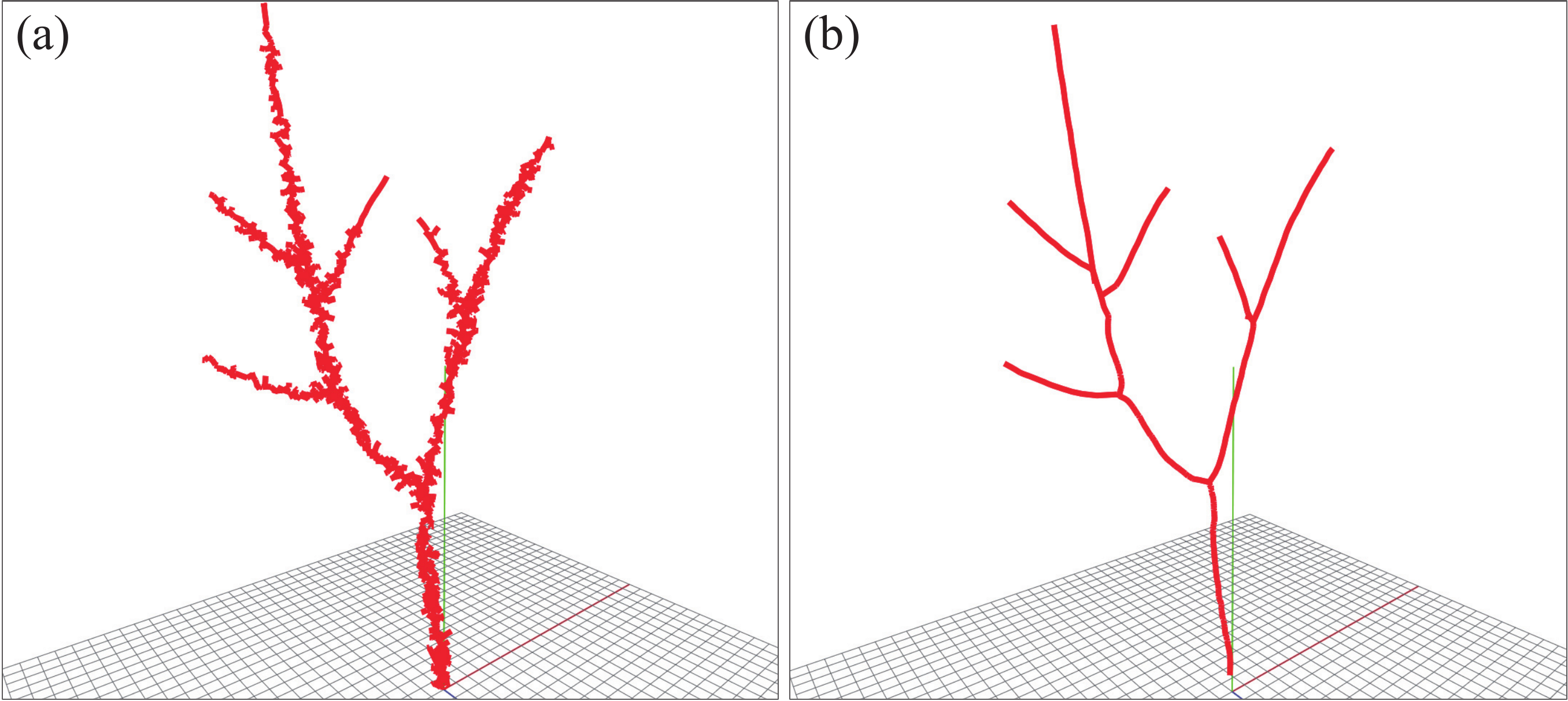}
\end{center}
\vspace{-3mm}
   \caption{Branch structure refinement: a) Branch structure candidates by flow simulation, and b) the final output after refinement and simplification.}
\label{fig:simplification}
\end{figure}

While the flow simulation generates a lot of branch candidates as shown in \fref{fig:simplification}(a), they are simplified and refined to yield the final structure. The refinement process involves 
(1) Smoothing: Apply low-pass filter along branch paths. We simply update the position of each vertex to the mass center of neighboring vertices, (2) Refinement: Move each vertex toward the direction of local probability maximum perpendicular to current branch direction, and (3) Simplification: Unify vertices located close to each other, and delete subtrees that locate in areas with small probability. 
We iterate the steps (1)--(3) several times and acquire the final branches as a tree graph structure as shown in \fref{fig:simplification}(b).

\subsection{Implementation Details}
\label{subsec:implementation_details}
Here we describe some implementation details on the Bayesian image-to-image translation and 3D branch generation.

\paragraph{Generator training.}
For generating the view-wise probability maps $\{B_{2D_{i}}\}$, we train the Pix2Pix network using images rendered using $10$ synthetic plants. The plant models are created by changing the parameters of a self-organizing tree model~\cite{palubicki09}, which is implemented in L-studio~\cite{lstudio}, an L-system-based plant modeler. For each plant, we render images viewed from $72$ viewpoints, and each image is additionally flipped for data augmentation. As a result, the total number of images is $10 \times 72 \times 2 = 1440$. Because Pix2Pix employs PatchGAN~\cite{isola17} that divides an image into patches for training, we find that this number of images is sufficient for training the generator $\pi_i$ of \eref{eq:2D_branch_probability}. 

The 2D branch probability estimation is implemented by modifying a TensorFlow implementation of Pix2Pix~\cite{pix2pix_tf}. To realize Bayesian image-to-image translation described in~\sref{subsec:bayesian_branch_probability_generation}, we modify the original implementation by adding dropout layers to each of four central layers of the encoder and decoder blocks. Since we intend to model the uncertainty of contextual information in deeper layers (\ie, branch patterns), we set a larger dropout rate in central layers ($0.8$ for the innermost layer, $0.7$ and $0.6$ for the next two, and $0.5$ for the outermost layer). Although~\cite{kendall17} uses a fixed dropout rate ($0.5$), they also report that applying dropout only for central layers yields better performance for segmentation tasks. We additionally use dropout for all the skip connections with a fixed dropout rate $0.3$. The inference is repeated $100$ times by randomly changing the dropout patterns. A previous work reports tens of iterations are sufficient to obtain the probability distributions~\cite{kendall17}. 

\vspace{-3mm}
\paragraph{Explicit branch generation.}
For particle flow simulation described in \sref{subsec:paticle_flow}, we generate approximately $10,000$ particles in the 3D space. 
Let $\log B_{3D}(\mathcal{S}_r(\vp_{t}))$ as the subset of the log-probability map around the particle location $\vp_{t}$ within radius $r$, we computed $\vF_c(\vp_{t})$ and $\vF_d(\vp_{t})$ as the mass center and a major axis of $\log B_{3D}(\mathcal{S}_r(\vp_{t}))$. 
We set $\lambda_r$ as a constant ($0.1$ in our experiment) but change $\lambda_c$ and $\lambda_d$ according to the distance $d_c$ between $\vp_{t}$ and mass center of $B_{3D}(\mathcal{S}_r(\vp_{t}))$ as
\begin{align}
\lambda_c = \frac{d_c}{r}(1-\lambda_r), \quad \lambda_d = 1-\lambda_r-\lambda_c.
\end{align}
In this manner, $\lambda_c$ becomes smaller when the mass center gets closer to $\vp_{t}$. The implementation of the weight control is based on the previous approach~\cite{neubert07}.

\vspace{-3mm}
\paragraph{Resolution and processing time.}
We used $128 \times 128$ [px] images for Pix2Pix, and they were upsampled to $1024\times 1024$ [px] for the 3D aggregation process. 
With our unoptimized implementation, it took approximately $10$ [sec] for the Pix2Pix with variational inference ($100$ inferences) per a viewpoint using an NVIDIA GeForce GTX 1080 GPU. Also, it took $1$ [min] for 3D aggregation, and $30$ [sec] for particle simulation on a CPU (3.70GHz, 6 cores). 


\begin{table*}
\small
\begin{center}
\begin{tabular}{|r|l||c|c|c|c||c|c|c|c|}
\hline
 \multicolumn{2}{|c||}{} & \multicolumn{4}{|c||}{Geometric error} & \multicolumn{4}{|c|}{Structure error} \\
 \multicolumn{2}{|c||}{} & \multicolumn{4}{|c||}{(Euclidean distance)} & \multicolumn{4}{|c|}{(difference in joint numbers)} \\ \hline
&Number of cameras 
& 72			& 36			& 12			& 6				& 72		& 36		& 12		& 6		 \\ \hline\hline
\multirow{4}{15mm}{\centering2D\\branch map\\$\{B_{2D_i}\}$} 
&Visible branch region
& 4.20			& 3.98			& 3.59			& 18.05 		& 0.6		& {\bf 0.6}	& {\bf 1.0}	& 40.4 \\ \cline{2-10}
&Whole plant region~\cite{neubert07}
& 12.37			& 18.12			& 24.75			& 17.44  		& 305.4		& 117.6		& 65.4		& {\bf 35.0}	 \\ \cline{2-10}
&Image-to-image translation	
& 1.76			& 2.26			& 2.40			& 14.95  		& 0.8		& 8.2		& 13.6		& 88.2	 \\ \cline{2-10}
&Bayesian image-to-image translation (proposed)
& {\bf 1.69}	& {\bf 1.74}	& {\bf 2.14}	& {\bf 14.53}  	& {\bf 0.4}	& 1.4		& 7.4		& 60.2	 \\ \hline
\end{tabular}
\end{center}
\vspace{-5mm}
\caption{The accuracy of 3D branch structure generated using different settings and number of views (averaged over five simulated plants). The geometric error is defined on a relative scale.}
\vspace{-4.5mm}
\label{tab:results_cg}
\end{table*}

\begin{figure}[t]
\begin{center}
\includegraphics[width=\linewidth]{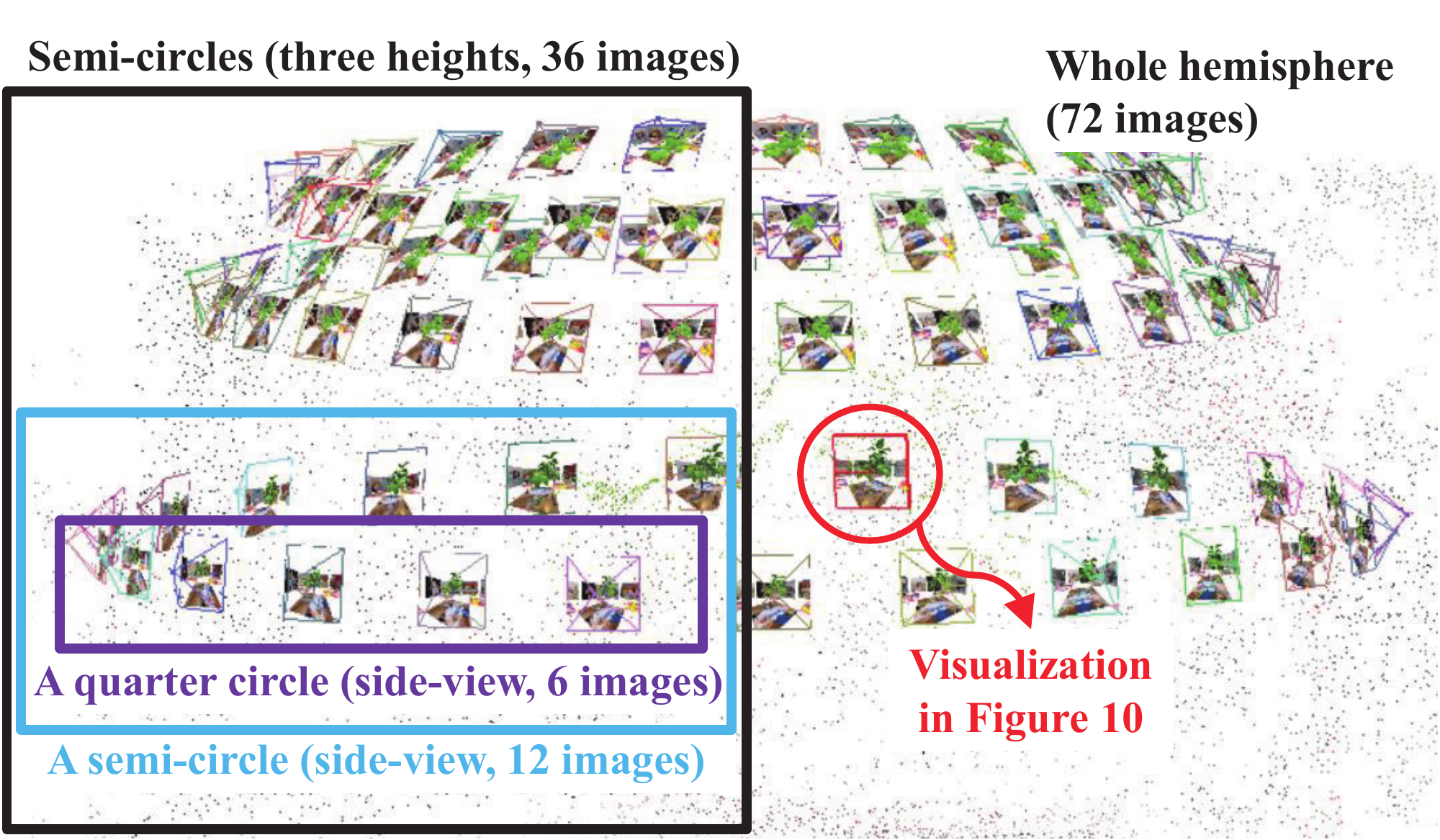}
\end{center}
\vspace{-5mm}
   \caption{Camera setting employed in the experiments.} 
\label{fig:cameras}
\vspace{-2mm}
\end{figure}

\section{Experiments}
\label{sec:experiments}
\vspace{-1mm}
We conduct experiments using simulated and real-world plant images and assess the quality of the reconstruction both quantitatively and qualitatively.

\subsection{Simulated plants}
\vspace{-1mm}
\label{sec:simulation}
\paragraph{Quantitative evaluation.}
We first describe the experimental result using simulated plant models, which have the ground truth branches for quantitative evaluation. The accuracy of generated plants by the proposed method is assessed using two metrics; \emph{geometric} and \emph{structure} errors.

The geometric error evaluates the Euclidean distances between 3D points in the generated 3D branch structure and points in the ground truth branches. We use a graph representation of the generated 3D branch structure and the ground truth. By sampling 3D points on the graph edges of both generated and the ground truth branches, we assess the geometric error. Let $\vg \in \mathcal{G}$ and $\vt \in \mathcal{T}$ be generated and true 3D branch points, respectively. The geometric error is defined as a bidirectional Euclidean distance $d$~\cite{zhu12} between the two point sets written as
\begin{align}
\small
d(\mathcal{G}, \mathcal{T}) = \frac{1}{2} \left( \frac{\sum_\mathcal{G} ||\vg - N_{\mathcal{T}} (\vg)|| }{|\mathcal{G}|} + \frac{ \sum_{\mathcal{T}} || \vt - N_{\mathcal{G}} (\vt)|| }{|\mathcal{T}|} \right), \nonumber
\end{align}
where $N_{\mathcal{G}}(\vx)$ and $N_{\mathcal{T}}(\vx)$ are functions to acquire the nearest neighbor point to $\vx$ from point sets $\mathcal{G}$ and $\mathcal{T}$, respectively, and $|\mathcal{G}|$ and $|\mathcal{T}|$ denote the numbers of points in $\mathcal{G}$ and $\mathcal{T}$. The geometric error is defined only up to scale because our branch recovery method is also up to scale like most multi-view 3D reconstruction methods.

\begin{figure*}[t]
\begin{center}
\includegraphics[width=\linewidth]{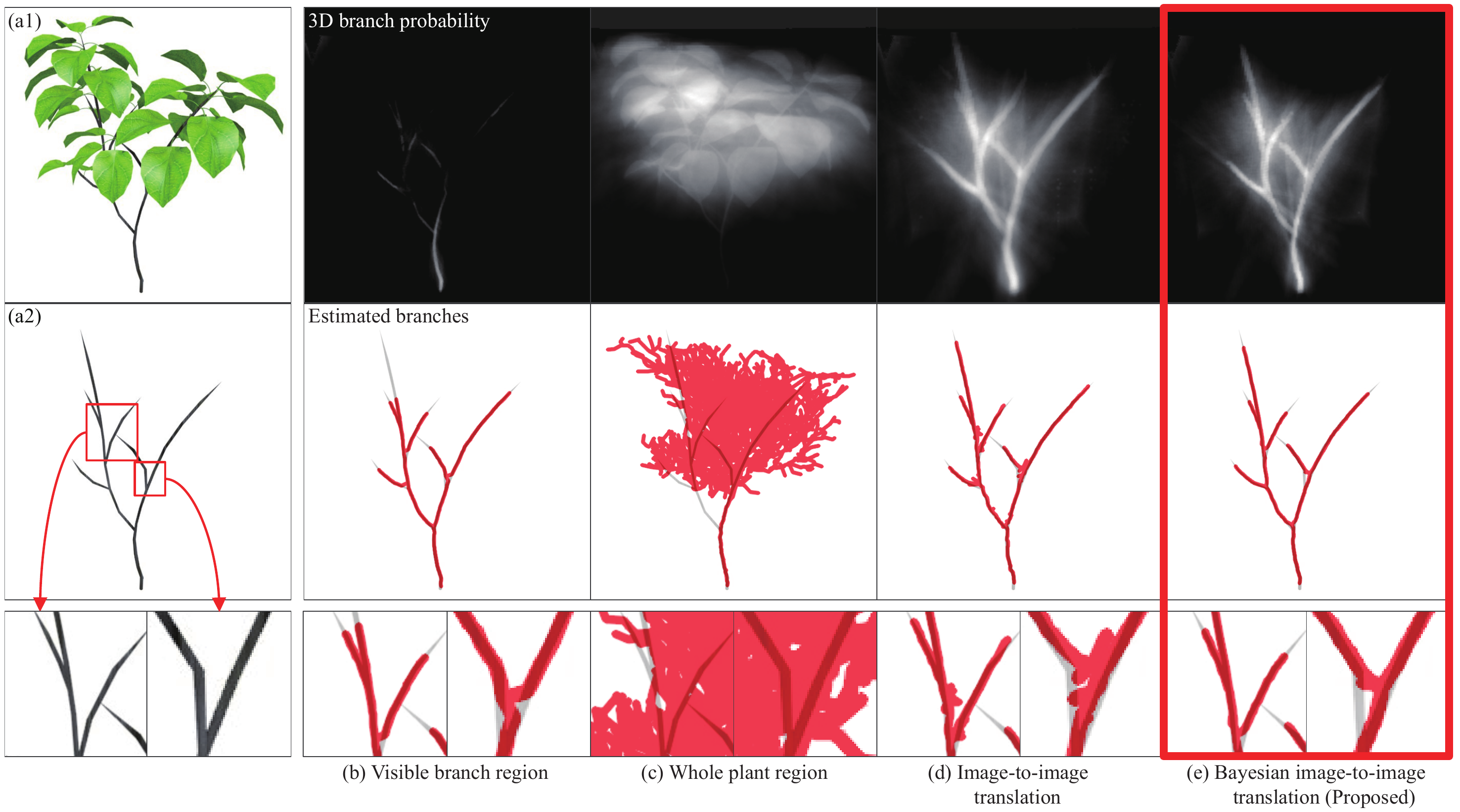}
\end{center}
\vspace{-6mm}
   \caption{Results using a simulated plant. The leftmost column shows (a1) an input image and (a2) the ground truth branch structure. 
   The other columns show the 3D branch probability and the resultant branch structures using different settings. 
   In comparison to other settings, the proposed approach (e) generates accurate and stable branch structure.}
\label{fig:results_cg}
\vspace{-3mm}
\end{figure*}

\begin{figure}[t]
\begin{center}
\includegraphics[width=\linewidth]{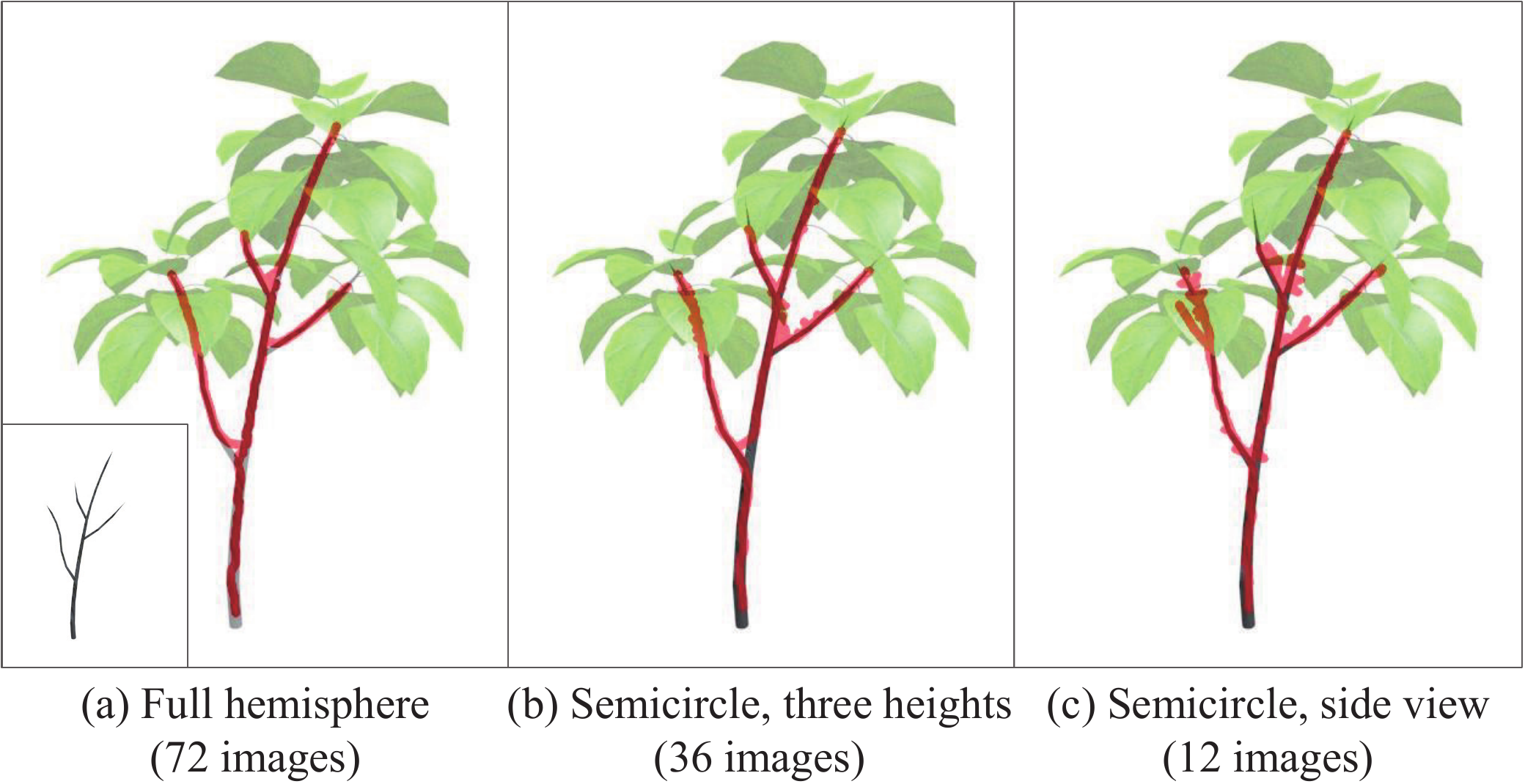}
\end{center}
\vspace{-3mm}
   \caption{Results when reducing the cameras. The ground truth branch is shown in bottom left corner.}
\label{fig:few_camera}
\vspace{-3mm}
\end{figure}

We define the structure error as the difference in joint numbers between the generated and the ground truth structures. 
From a tree graph generated by the explicit branch generation, we count the number of graph vertices, where the number of connected edges is three (\ie, one incoming and two outgoing) or more, for assessing the accuracy of structure recovery.

Using these error metrics, we assess the method using four different settings described below:
\begin{itemize}
\vspace{-0.5mm}
\setlength{\parsep}{0pt}
\setlength{\parskip}{0pt}
\item {\bf Visible branch region}: As the easiest test case, we use the ground truth branch regions in multi-view images that are not occluded by leaves. It naturally skips the image-to-image translation process because the 2D branch structure map $\{B_{{2D}_i}\}$ is directly given.
\item {\bf Whole plant region}: In this test case, to compare with the image-to-image translation strategy in our problem, a whole plant (\ie, branches and leaves) region is directly used without image-to-image translation for defining the probability map $\{B_{{2D}_i}\}$. This setting is akin to the previous tree modeling method~\cite{neubert07}, which generates branches that fit the entire volume of a tree by a particle flow simulation. 
\item {\bf Image-to-image translation}: In this case, the 2D probability map $\{B_{{2D}_i}\}$ are generated by the original image-to-image translation~\cite{isola17}, where the Bayesian extension is not employed. 
\item {\bf Bayesian image-to-image translation} {\it(Proposed)}: In this setting, we use the proposed method, with which branch probability maps $\{B_{{2D}_i}\}$ are generated by the Bayesian image-to-image translation.
\vspace{-0.5mm}
\end{itemize}

For the evaluation, we use five plant 3D models, where the branching parameters are different from the training dataset described in \sref{subsec:implementation_details}. To assess the effect of the varying number of cameras, we generate branch structures from multi-view images (a) covering a whole hemisphere ($72$ cameras), (b) camera paths on semi-circles at three heights ($36$ cameras), (c) a semi-circle ($12$ cameras), and (d) a quarter circle ($6$ cameras), as illustrated in \fref{fig:cameras}.

\vspace{-2.5mm}
\paragraph{Results.}
\Tref{tab:results_cg} summarizes the accuracy evaluation of the four settings with varying number of views, and \fref{fig:results_cg} shows some of the results when using $72$ images.
The proposed approach generates accurate branch structures across all the view settings in geometric error, while the accuracy degrades in the smaller number of views. 
For the case of ``visible branch regions,'' although we extracted the ground truth branch labels seen from each camera, branches are occluded by leaves and there are parts that are unobserved from most of the viewpoints, resulting in greater geometric errors.
The case of ``image-to-image translation,'' the output includes branch paths that are not supported by the reliability of inference, instead generated from one sample, which results in small sub-branches (see \fref{fig:results_cg}(d)). Since the proposed approach averages multiple inferences, the estimated branches become more stable and precise compared to the non-Bayesian approach, thus shows better agreement between the estimated probability map and the branch paths.

\Fref{fig:few_camera} compares the results of the proposed method together with the Bayesian image-to-image translation along the varying number of views. It can be seen that the more views make the estimation more faithful, while the method is still able to recover the overall branch structures.

\begin{figure}[t]
\begin{center}
\includegraphics[width=\linewidth]{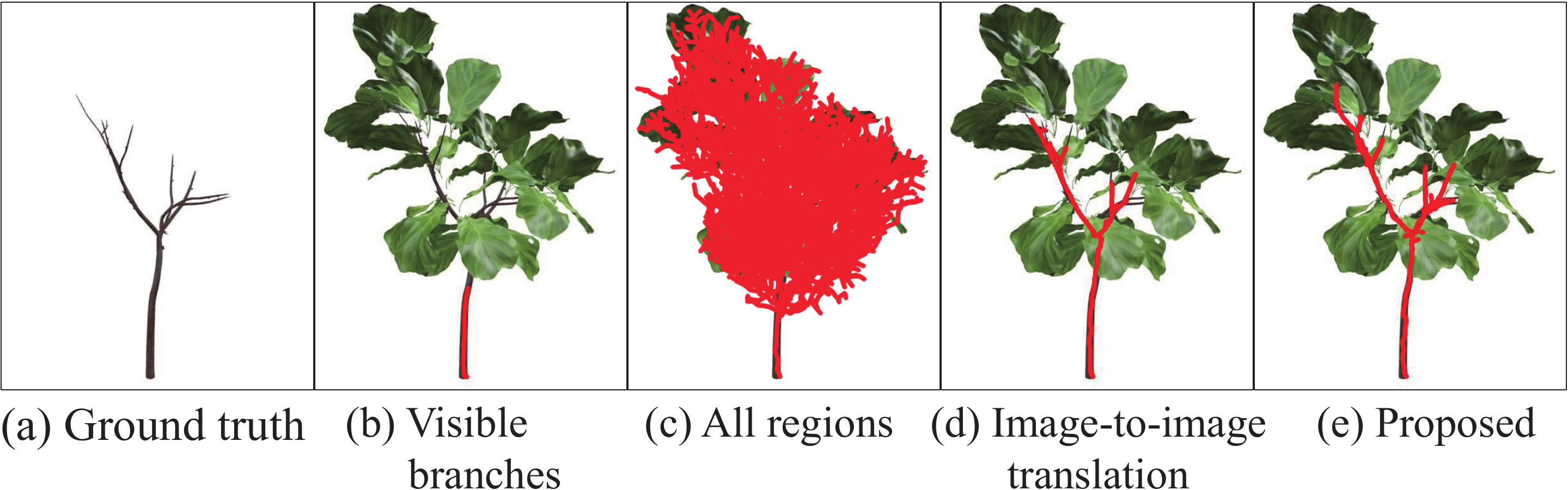}
\end{center}
\vspace{-3.55mm}
   \caption{Results using a plant of a different species than the ones used for training.}
\label{fig:species}
\vspace{-3mm}
\end{figure}

\begin{figure*}[t]
\begin{center}
\includegraphics[width=\linewidth]{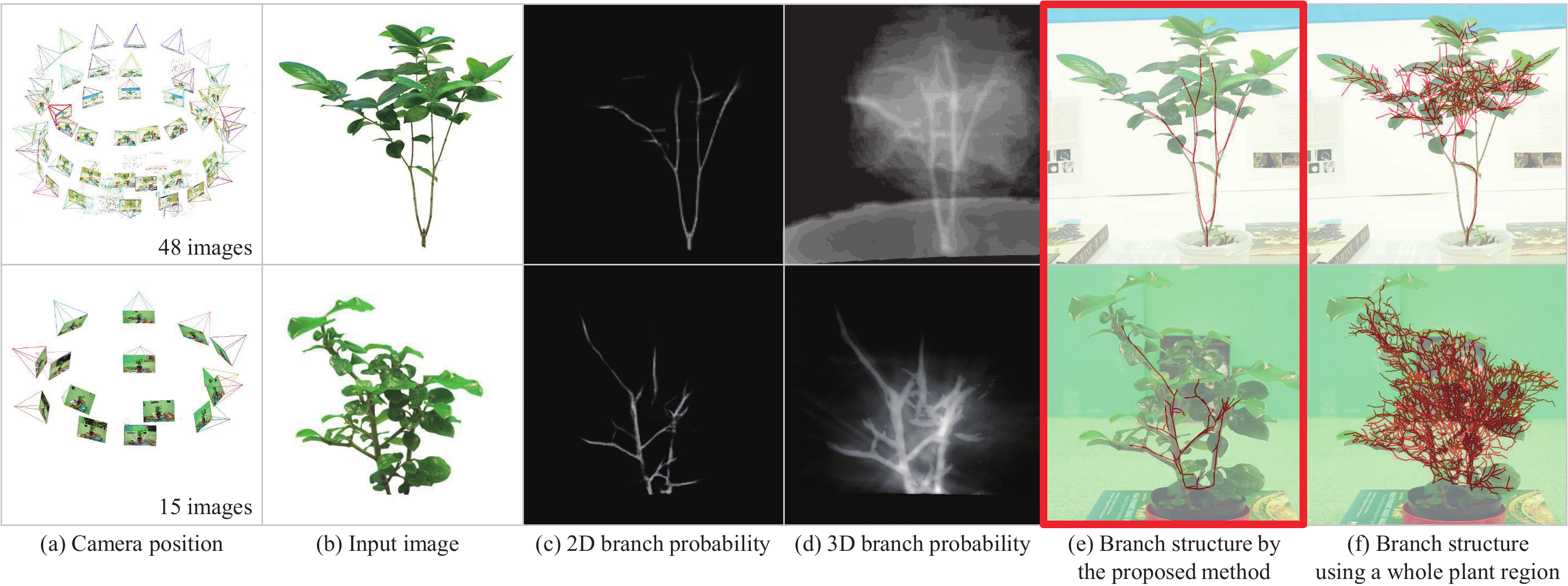}
\end{center}
\vspace{-3mm}
   \caption{Results using real plants. 
   Using input images where the foreground plants are extracted (b), the proposed approach generates convincing branch structures (e), in comparison to branch structures generated by a previous tree reconstruction approach~\cite{neubert07} (f).}
\label{fig:results_real}
\vspace{-2mm}
\end{figure*}

\vspace{-3mm}
\paragraph{Different species.}
To study the generalization ability, we apply the proposed method to plant models where their species (\ie, texture and shape of leaves) are different from the simulated plants used for training. \Fref{fig:species} shows that the proposed method still generates accurate branch structures compared to the other approaches even for different leaf textures.

\subsection{Real plants}
\vspace{-1.3mm}
Now we show the result of the proposed method using the images of real-world plants. 
In this experiment, we use the same trained model for the image-to-image translation that is used in the simulation experiment in~\sref{sec:simulation}. 
To avoid the unmodelled factors in the experiment, we first manually extract the plant regions from images, which can be alternatively achieved by chroma keying.
\Fref{fig:results_real} shows the results including the intermediate outputs in comparison to the method~\cite{neubert07}. 
The proposed method qualitatively yields convincing branch structures, even though the image-to-image translation network is trained using simulated plants. Compared with a previous tree reconstruction approach~\cite{neubert07}, our approach shows its effectiveness in generating the branch structure under leaves.

\begin{figure}[t]
\begin{center}
\includegraphics[width=0.9\linewidth]{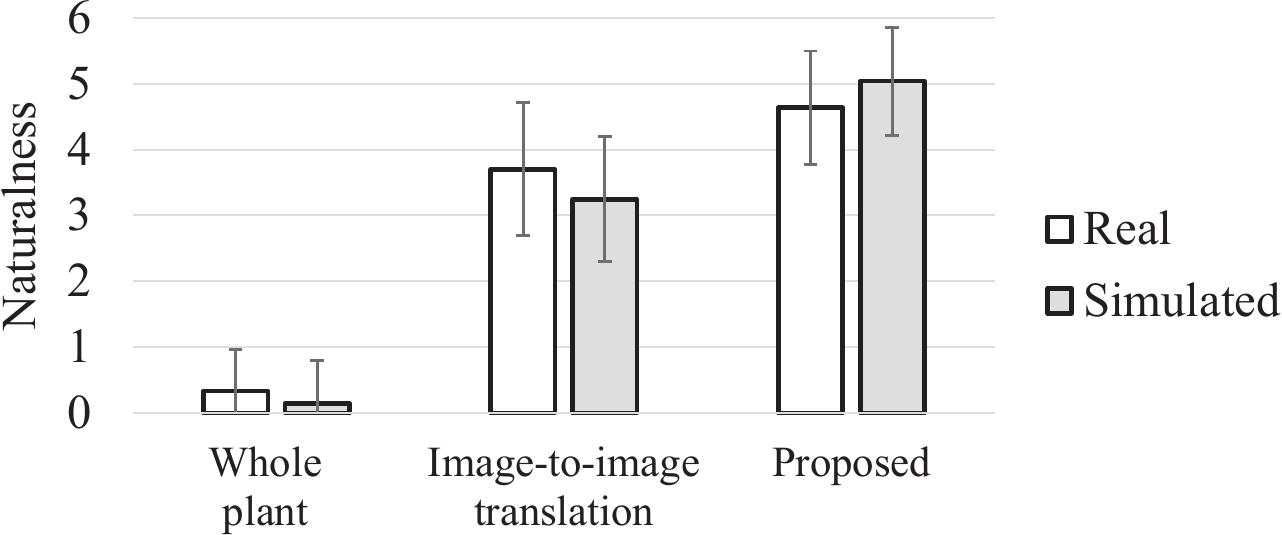}
\end{center}
\vspace{-4mm}
   \caption{Subjective evaluation of the branch naturalness.}
\vspace{-4mm}
\label{fig:comp}
\end{figure}

Since for this experiment we did not have access to the ground truth, we conducted a small subjective evaluation by $10$ participants to assess the perceptual naturalness of output plant skeletons. The participants were asked to watch the skeletons generated by each method overlaid on the plant image and assessed naturalness of the skeletons in a $7$-step Likert scale. \Fref{fig:comp} summarizes the result. The proposed approach yields the best score among the comparisons, and the trend is consistent with the evaluation of simulated plants. 

\vspace{-0.5mm}
\section{Discussion}
\vspace{-1.5mm}
We presented a plant modeling approach via image-to-image translation to estimate branch structures of 3D plants from multi-view images, even if the branches are occluded under leaves. 
The combination of Bayesian image-to-image (leafy- to branch-image) translation and 3D aggregation generates the branch existence probability in a 3D voxel space, resulting in a probabilistic model of a 3D plant structure. We have shown that explicit branch structures can be generated from the probabilistic representation via particle flow simulation.

The experimental results using simulated plants showed that the proposed approach is able to generate the accurate recovery of the branch structure of a plant compared with a previous tree modeling approach~\cite{neubert07}. It has also been shown that a Bayesian extension of image-to-image translation is effective in obtaining a stable estimate of the branch structure in the form of probability compared with a non-Bayesian one. Qualitatively, the result also applies to the real-world plants as demonstrated in the experiment.


One of the advantages of our probabilistic approach is that it can generate a variety of plant branch structures from a single branch probability map, by changing the parameters of particle flow simulation, as shown in \fref{fig:diff_structures}, which may benefit computer graphics applications. 


\vspace{-0.5mm}
\section*{Acknowledgement}
\vspace{-1.5mm}
This work was supported in part by JST PRESTO Grant Number JPMJPR17O3.

{\small
\bibliographystyle{ieee}
\bibliography{CVPR2018_isokane}
}

\end{document}